\documentclass{article}
\usepackage[utf8]{inputenc}
\usepackage{iclr2026_conference,times}

\usepackage{amsmath,amsfonts,bm}

\def\eqref#1{equation~\ref{#1}}

\def\1{\bm{1}}

\DeclareMathAlphabet{\mathsfit}{\encodingdefault}{\sfdefault}{m}{sl}
\SetMathAlphabet{\mathsfit}{bold}{\encodingdefault}{\sfdefault}{bx}{n}

\usepackage{graphicx}
\usepackage{hyperref}
\usepackage{url}
\usepackage{cleveref}
\usepackage{wrapfig}
\usepackage{listings}
\usepackage{xcolor}
\usepackage{enumitem}
\usepackage{tabularx}
\usepackage{array}
\colorlet{light-gray}{gray!30}

\title{Test-Time Adaptation for LLM Agents via Environment Interaction}

\iclrfinalcopy

\author{
Arthur Chen$^{1}$\thanks{Work done during internship. Correspondence to: \texttt{haonan.chen@uwaterloo.ca}} \quad
Zuxin Liu$^{2}$ \quad
Jianguo Zhang$^{2}$ \quad
Akshara Prabhakar$^{2}$ \quad
Zhiwei Liu$^{2}$ \\
\textbf{Shelby Heinecke}$^{2}$ \quad
\textbf{Silvio Savarese}$^{2}$ \quad
\textbf{Victor Zhong}$^{1}$\thanks{Equal advising.} \quad
\textbf{Caiming Xiong}$^{2}$\footnotemark[2] \\
\\
$^{1}$University of Waterloo \qquad $^{2}$Salesforce AI Research
}

\newcommand{\todo}[1]{\marginpar{\textcolor{red}{TODO}}}

\newcommand{\parametricmethodname}{syntactic alignment}
\newcommand{\nonparametricmethodname}{dynamics grounding}
\newcommand{\parametricmethodnameshort}{SA}
\newcommand{\nonparametricmethodnameshort}{DG}

\begin{document}

\maketitle
\begin{abstract}
Large language model (LLM)-based agents struggle to generalize to novel and complex environments, such as unseen websites or new sets of functions, due to a fundamental mismatch between their pre-training and test-time conditions.
This challenge stems from two distinct failure modes: a syntactic misunderstanding of environment-specific components like observation formats, and a semantic misunderstanding of state-transition dynamics, which are only revealed at test time.
To address these issues, we propose two distinct strategies for adapting LLM agents by leveraging environment-specific information from interaction that is available during deployment.
First, an \textit{online \parametricmethodname{}} (\parametricmethodnameshort) method parameterizes environmental nuances by learning a lightweight adaptation vector that biases the model's output distribution, enabling rapid alignment with an environment response format.
Second, a \textit{deployment-time \nonparametricmethodname{}} (\nonparametricmethodnameshort) method employs a persona-driven exploration phase to systematically probe and learn the environment's causal dynamics before task execution, equipping the agent with an in-context world model.
We evaluate these strategies across diverse agentic benchmarks, including function calling and web navigation.
Our empirical results show the effectiveness of both strategies across all benchmarks with minimal computational cost.
We find that dynamics grounding is particularly effective in complex environments where unpredictable dynamics pose a major obstacle, demonstrating a robust path toward more generalizable and capable LLM-based agents.
For example, on the WebArena multi-site split, this method increases the agent's success rate from 2\% to 23\%.
We release our code\footnote{\url{https://github.com/r2llab/GTTA}}.
\end{abstract}

\section{Introduction}
Large language model (LLM)-based AI agents have demonstrated remarkable capabilities across diverse tasks~\citep{li_can_2023, zhou_webarena_2024, patil_berkeley_nodate, xie_osworld_2024}, yet they often fail to generalize when deployed in novel agentic environments such as web navigation~\citep{zhou_webarena_2024, he_webvoyager_2024} and function calling~\citep{patil_berkeley_nodate, yao_tau-bench_2024}.
These failures reflect two core challenges---the systematic mismatch between the agent's pre-trained knowledge and the specific syntax and dynamics of the deployed environment~\citep{yang_agentoccam_2025, thil_navigating_2024}. Although both mismatches stem from a lack of prior knowledge, we distinguish them operationally because they require different adaptation mechanisms (syntactic vs. semantic). To illustrate these two failure patterns, we consider an LLM agent tasked with booking a flight on a previously unseen travel website:
\begin{enumerate}[leftmargin=*]
    \item \textbf{Syntactic Mismatch}:
    the LLM agent's prior knowledge does not align with environment-specific information, such as observation structure~\citep{yang_if_2024,gur_real-world_2023} and the environment's unique syntax~\citep{lei_spider_2024, chen_agent-flan_2024}.
    This causes parsing and context-understanding issues: a pre-trained LLM agent might try \texttt{click(``Search'')} or target \texttt{destination}, while the new site exposes \texttt{Go} and \texttt{dest\_field}, producing invalid actions and immediate failures.~\citep{yang_if_2024,gur_real-world_2023,lei_spider_2024,chen_agent-flan_2024}

    \item \textbf{Semantic Mismatch}: the agent lacks an accurate, environment-specific causal model of state transitions, so it cannot predict the consequences of actions. For example, the agent may expect \texttt{click(``Go'')} to show flight results, but the site instead opens a date-confirmation pop-up; without knowing this transition, the agent cannot form the correct multi-step plan and fails.~\citep{forrester_counterintuitive_1971,chae_web_2025,zhang_attacking_2024}
\end{enumerate}
Current approaches to this generalization gap are ill-suited for rapid adaptation in novel environments.
In addition, many methods require human-annotated or LLM-annotated demonstrations~\citep{wang_agent_2024, murty_nnetnav_2025,xu_agenttrek_2024}, which can be resource-intensive or rely on LLM's prior knowledge about the environment.
On the other hand, to address semantic mismatch, explicit world modeling like~\citep{chae_web_2025} involves a heavyweight pipeline of extensive data collection and fine-tuning a separate, specialized model that is computationally expensive and struggles to generalize without being retrained.
Both paradigms present significant overhead, highlighting the need for more efficient strategies that can ground an agent using only the information available at test time.

To address the two failure modes and close the generalization gap, we propose two annotation-free strategies that operate at deployment time.
(1) Online \parametricmethodname{} (\parametricmethodnameshort)---a lightweight module that learns a small per-interaction bias (an adaptation vector applied to late features) to quickly align an agent’s output distribution with environment-specific syntax. Observing \texttt{Go} and \texttt{dest\_field} on the page, for example, lets the adapter steer action generation toward the site’s actual element names.
(2) Dynamics grounding (\nonparametricmethodnameshort)---a short, persona-guided exploration phase that iteratively collects in-context rules about state transitions (e.g., clicking \texttt{Go} opens a date pop-up) and supplies those rules as context for downstream planning. Both methods require only test-time observations and no trajectory annotations or expensive fine-tuning.

We evaluate our methods on function-calling and web-navigation benchmarks. Both approaches yield consistent improvements: \parametricmethodname{} gives efficient, steady gains across sites, while dynamics grounding produces particularly large improvements on environments with unfamiliar transitions. For example, on the WebArena multi-site split~\citep{zhou_webarena_2024}, dynamics grounding raises \texttt{GPT-4.1} success from 2\% to 23\%. Taken together, syntactic and semantic test-time grounding substantially narrow the deployment generalization gap and offer a practical path toward more robust LLM agents.


\section{Related Work}

\paragraph{Test-Time Adaptation}
Originating in computer vision, test-time adaptation (TTA) addresses distributional shifts between training and testing data by adapting models at inference time~\citep{wang_tent_2021,niu_efficient_2022,sun_test-time_2020}.
Our \parametricmethodname{} builds directly on methods that update model parameters or steering vectors at test time to minimize an unsupervised objective, such as entropy minimization or self-supervised learning~\citep{wang_tent_2021,hu_slot_2025,li_inference-time_2023}.
Our \parametricmethodname{} aligns with steering vector research~\citep{tan_analyzing_2025, sinii_steering_2025, panickssery_steering_2024, turner_steering_2024} in its use of latent space biases to shift output distributions. However, standard steering approaches generally apply modulation vectors to shift high-level traits ~\citep{tan_analyzing_2025}, such as honesty~\citep{zou_representation_2025}. We distinguish our work by tailoring this mechanism to the dynamic nature of agentic tasks: our vector is treated as a temporary parameter, updated online using the environment's response as a self-supervisory signal, and reset per episode to ensure precise, context-dependent alignment without interference across tasks.
While TTA for LLMs is emerging for tasks like math reasoning~\citep{zuo_ttrl_2025} and few-shot learning~\citep{akyurek_surprising_2025}, its application to the unique, interactive, and stateful challenges of LLM agents remains underexplored.
Our work is the first to systematically apply this paradigm to adapt agents to novel environment formats without supervised trajectories.

\paragraph{Environment Modeling for Agents}
Recent work has focused on incorporating environment dynamics into LLM-based planning by training a parameterized world model~\citep{fang_webevolver_2025, chae_web_2025, ding_understanding_2026, qiao_agent_2024}, or using an LLM to predict the next state using continuously updated environment rules~\citep{zhou_wall-e_2024}.
Such approaches typically require collecting a large set of observations for learning a world model, and can be computationally expensive~\citep{chae_web_2025}.
In contrast, our \nonparametricmethodname{} introduces a lightweight, deployment-time pipeline to generate a temporary, in-context world model from a few exploratory interactions.
This aligns with a broader trend of using in-context learning for adaptation, including methods like agent workflow memory~\citep{wang_agent_2024} and retrieval-augmented generation (RAG) for planning~\citep{luo_reasoning_2023, kagaya_rap_2024, wang_learning_2024}.
However, our contribution lies in the systematic and automated process of generating this knowledge via persona-driven exploration, removing the need for pre-existing annotated demonstrations or a separate, trained dynamics model.

\paragraph{LLM-Based Agents}
Large language models (LLMs) have become the planning backbone of a wide range of agentic systems due to their strong instruction-following capabilities and generalization across diverse tasks~\citep{ouyang_training_2022, mishra_cross-task_2022, zhou_webarena_2024}.
Their ability to interpret complex instructions and generate coherent action plans has enabled their integration into applications such as web navigation~\citep{zhou_webarena_2024, he_webvoyager_2024, lu_weblinx_2024, deng_mind2web_2023} and function calling~\citep{patil_berkeley_nodate, yao_tau-bench_2024}.
Despite recent advances, LLM agents struggle to comprehend the specific syntactical input structure of different environments, such as in web navigation~\citep{yang_if_2024,gur_real-world_2023} and text-to-SQL~\citep{lei_spider_2024}.
This is due to the fundamental mismatch between the LLM pre-training corpus and the specific agentic environment~\citep{chen_agent-flan_2024, ben-david_theory_2010}.


\section{Grounded Test-Time Adaptation Strategies}

In this section, we detail our two adaptation strategies which mitigate syntactic and semantic gaps.
We first formalize the problem setup, which is designed to reflect realistic deployment scenarios and is defined by three key constraints:
\begin{enumerate}[leftmargin=*]
\item \textbf{No annotated trajectories or offline data.} The agent cannot access expert demonstrations or pre-collected data from the target environment.
It must learn from scratch during interaction.
For our running travel example, the agent has never seen travel-example.com and has no pre-existing dataset of successful bookings on that site.

\item \textbf{Online, streaming adaptation.} The agent receives tasks one at a time and must adapt on-the-fly without seeing a full batch of test instances in advance.
For example, it must adapt to the website's quirks while handling a single request (e.g.~``Book a flight from New York to Saskatoon'') before seeing any future tasks.

\item \textbf{Permitted test-time interaction.} The agent is allowed to interact with the environment to gather information before receiving a specific task, but without supervision.
For instance, it can perform a ``blind'' exploration of travel-example.com to discover that a button triggers a calendar pop-up, but it does not know what kind of tasks the user will ultimately ask it to execute.
\end{enumerate}

This setting mirrors constraints in practical benchmarks (e.g., BIRD~\citep{li_can_2023}, WebArena~\citep{zhou_webarena_2024}) and necessitates methods that can leverage only unsupervised, test-time interaction.
To formulate this challenge, at test-time the LLM input $\mathcal{I}$ is constructed as
\begin{equation}
    \mathcal{I} = [p; o; \{a\}_{i=1}^{T-1}]
    \label{eq:1}
\end{equation}
where $p$ denotes the task instruction (e.g., rules, task description), $o$ denotes the current environment observation (e.g., accessibility tree observation, response of function call), and $\{a\}_{i=1}^{T-1}$ denotes the sequence of past actions taken by the LLM-based agent up to the current step.
For environments with short or simple observations (e.g., tool-use scenarios), we instead construct the model input using all previous interactions between the agent and the environment, which may include both environment responses and agent actions.
With this formulation of the challenge in mind, we describe two adaptation methods in the remainder of this section.

\subsection{Syntactic Alignment}
\begin{figure}
   \vspace{-2em}
   \includegraphics[width=\linewidth]{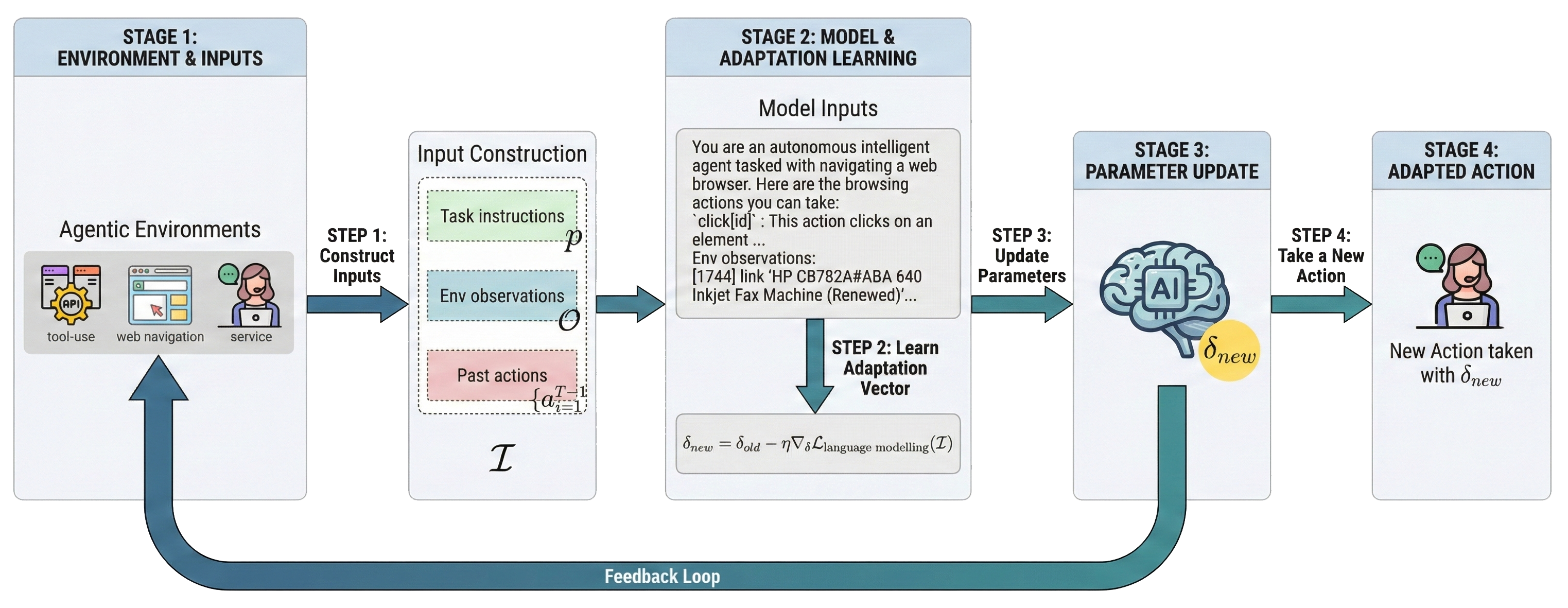}
   \caption{
   \textbf{Overview of \parametricmethodname\ (\parametricmethodnameshort).} This figure includes an example of web navigation shopping task to illustrate how the agent adapts to new environment.
   (1) At the start of each episode, we initialize an adaptation vector $\delta$ as a zero vector and construct inputs to the LLM agent.
   (2) During task execution, the agent receives environment instructions and observations.
   (3) At each step, we update the adaptation vector using cross-entropy loss on the current input, and apply the adaptation vector as a bias to the LLM's final hidden layer. This enables rapid alignment to environment-specific observation and action formats.
   (4) The LLM agent takes a new action with the updated vector, which shifts the model's output distribution to better match the test-time environment.
   }
   \label{fig:parametric_overview}
\end{figure}

To address syntactic mismatches that arise in novel environments, we propose \textit{\parametricmethodname{}}.
The objective is to parametrically adapt the model's output distribution at test time by treating the current context (i.e.~task instructions, observations, and action history) as a self-supervisory signal.
This allows the model to align with local syntactic patterns, such as unique UI element labels or response formats, without requiring a deep understanding of the environment's causal dynamics.

Our approach introduces a single lightweight adaptation vector $\delta \in \mathbb{R}^d$, where $d$ is the hidden dimension of the language model.
This vector acts as an additive bias to the final hidden representations $H \in \mathbb{R}^{n \times d}$ before the final projection layer.
At each step of an episode, the adapted logits are computed as:
\begin{equation}
    \text{logits}' = (H + \delta) W_{\mathrm{LM}}^T
    \label{eq:2}
\end{equation}
where $W_{\mathrm{LM}} \in \mathbb{R}^{|V| \times d}$ is the model's output projection matrix and $|V|$ is the vocabulary size.

This vector $\delta$ is updated at each turn by performing one gradient descent step to minimize the language modeling loss of the current input context $\mathcal{I}$ according to the following:
\begin{equation}
    \delta_{\text{new}} \leftarrow \delta_{\text{old}} - \eta \nabla_{\delta} \mathcal{L}_{\text{CE}}\left(f_{\theta, \delta}(\mathcal{I}_{1:n-1}), \; \mathcal{I}_{2:n}\right)
    \label{eq:3}
\end{equation}
where $f_{\theta, \delta}$ is the LLM parameterized by its fixed weights $\theta$ and the adaptable vector $\delta$, and $\eta$ is the learning rate.
$\mathcal{I}_{1:n-1}$ denotes the input subsequence consisting of tokens $1$ through $n-1$, and $\mathcal{I}_{2:n}$ denotes the corresponding next-token targets obtained by shifting the sequence by one position. The loss in Eq.~\ref{eq:3} is therefore computed by predicting each token in $\mathcal{I}_{2:n}$ from its preceding context $\mathcal{I}_{1:n-1}$.
This update encourages the model to internalize the syntactic features present in the immediate context by adapting its output distribution to the environment context.

To illustrate this in practice, consider our agent on the unseen website \texttt{travel-example.com}.
The agent's input context $\mathcal{I}$ contains the accessibility tree, which includes strings such as \texttt{[145] <button>Go} and \texttt{dest\_field}.
\begin{enumerate}[leftmargin=*]
    \item \textbf{Initialization}: At the episode's start, the adaptation vector is initialized as a zero vector, $\delta \leftarrow 0$.
    \item \textbf{Prediction}: Without adaptation, the model's logits would likely favor the action \texttt{click(``Search'')}, based on its pre-trained priors.
    \item \textbf{Adaptation}: We compute the language modeling loss on the context $\mathcal{I}$ (Eq.~\ref{eq:3}). Because the strings \texttt{Go} and \texttt{\texttt{dest\_field}} are present in $\mathcal{I}$, the gradient update to $\delta$ will shift its values to favor their tokens.
    \item \textbf{Correction}: In the subsequent generation step, the adapted logits (Eq.~\ref{eq:2}) now assign a higher probability to the syntactically correct action, \texttt{click(``Go'')}.
\end{enumerate}
This process is computationally efficient, and steps 2-4 are repeated at every model prediction within an episode.
To prevent catastrophic forgetting and ensure that adaptation is specific to the current environment, $\delta$ is reset to zero at the beginning of each new episode.
This lightweight update aligns the model's pre-trained distribution with the specific distribution of the test-time environment.

\subsection{Dynamics Grounding}

\begin{figure}[h]
    \centering
    \vspace{-1em}
    \includegraphics[width=1\linewidth]{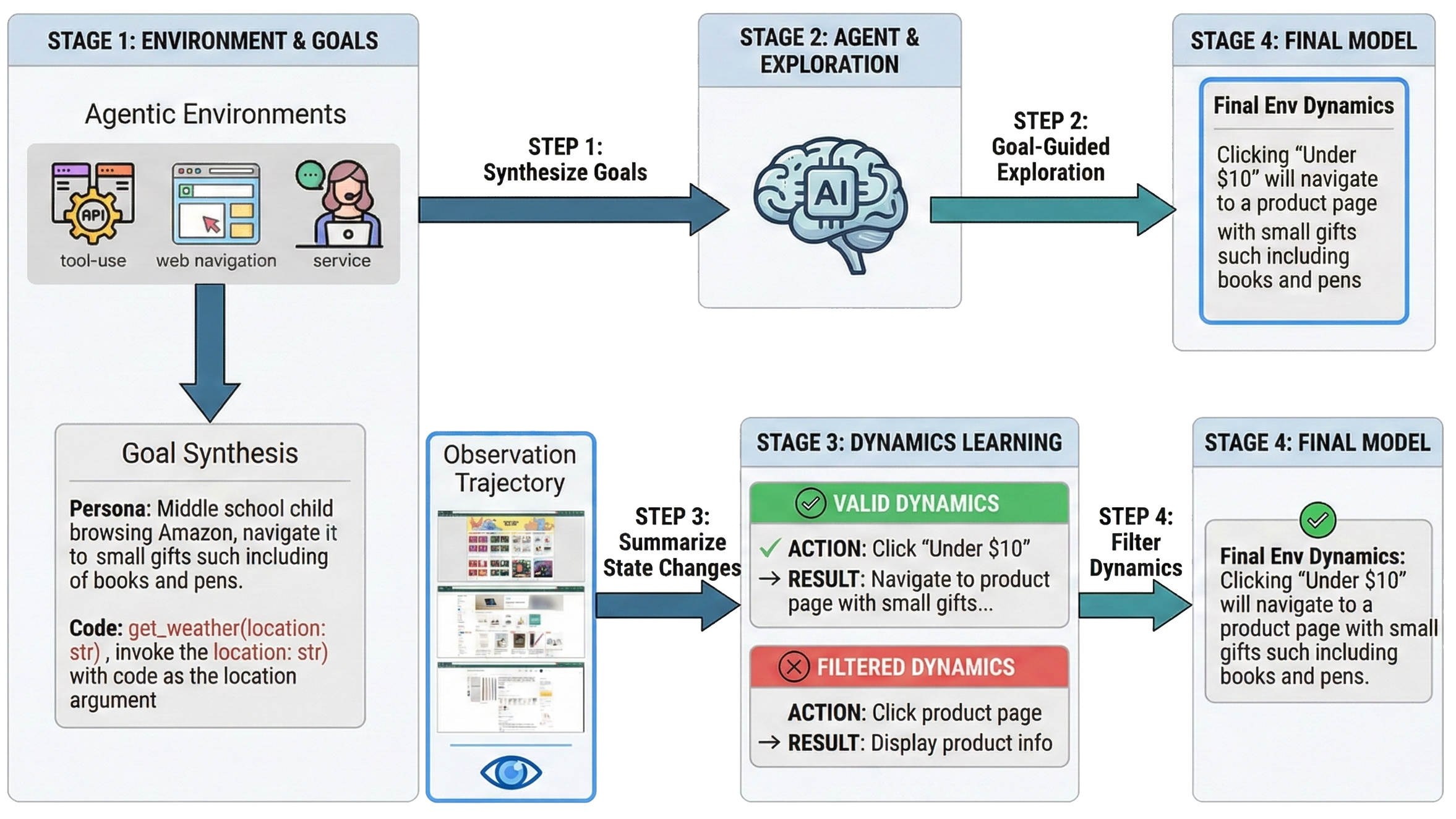}
    \vspace{-1em}
    \caption{
    \textbf{Overview of \nonparametricmethodname\ (\nonparametricmethodnameshort).} This figure includes an example of web navigation shopping task to illustrate how the pipeline generates environment dynamics in language. 
    (1) We synthesize diverse exploration tasks based on personas using environment descriptions.
    (2) An exploration agent interacts with the environment to collect interaction logs of state transitions.
    (3) An LLM extracts and summarizes environment dynamics from these logs.
    (4) A reasoning model filters less informative rules, which are then used to augment the agent's context during evaluation, enabling more transition-aware decision making.
    }
    \label{fig:nonparametric_overview}
    \vspace{-1em}
\end{figure}

LLM-based agents often fail in novel environments due to a \textbf{semantic mismatch}: they lack a ``world model'' to predict the outcomes of their actions~\citep{ha_world_2018, chae_web_2025}.
To address this, we propose \textit{\nonparametricmethodname}, a strategy that proactively discovers the causal dynamics of an environment \textbf{before the agent encounters any user tasks during deployment}.
The objective is to construct an in-context world model using natural language by performing a one-time, automated exploration.
The cost of this discovery phase is amortized over all subsequent tasks performed in that environment.

Our method follows a four-step pipeline to generate a concise set of environment dynamics, denoted as $E_{\text{clean}}$, for a given new environment.
\begin{enumerate}[leftmargin=*]
    \item \textbf{Persona/Exploratory Goal Synthesis.} An LLM is prompted with a high-level description of the environment (e.g., website purpose, API documentation) to generate a set of $N$ diverse, exploratory ``personas'' or ``goals''.
    Personas frame open-ended tasks designed to probe for non-obvious interactions.
    For our \texttt{travel-example.com} scenario, a synthesized persona might be: \textit{``As a first-time user, I want to see what happens if I try to search for a flight without selecting a date.''}.
    We prefer personas over a less-structured exploration policy (e.g.~maximizing state novelty or coverage) to guide the agent to explore complex, multi-step interactions that a naive exploration strategy might miss, thus yielding more semantically rich dynamics.
    \item \textbf{Exploration and on-the-fly Dynamics Extraction} For each persona or goal, an LLM agent interacts with the environment.
    The agent is instructed to take novel actions to maximize its discovery of new state transitions. After each action is executed and the system transitions into the next state, we take (observation, action, new observation) to summarize into a concise human-readable rule $e$. Immediately after, the list of generated rules $\{e\}_{i=1}^T$ is appended to the agent to encourage taking actions that have not been taken before.
    In our example, the agent would execute \texttt{click(``Go'')} and log the transition from the main page to a state where a calendar pop-up is active.
    Then the transition on \texttt{travel-example.com} would be extracted as the dynamic: Executing the action \texttt{click(`Go')} on the main page causes a calendar modal to appear for date selection.
    \item \textbf{Filtering and Consolidation.} The set of all extracted dynamics, $E=\{e\}_{i=1}^M$, is passed to a reasoning model.
    This model filters out trivial (e.g., ``typing in a text field shows text'') or repetitive rules, producing a final, clean set of dynamics, $E_{\text{clean}}$.
\end{enumerate}
During task execution at test time, we augment the input context $\mathcal{I}$ with discovered dynamics:
\begin{equation}
\mathcal{I}'=[\mathcal{I}; E_{\text{clean}}]   
\end{equation}
By leveraging the LLM's in-context learning capabilities, this explicit knowledge of the environment's state-transition patterns guides the agent to make more informed and transition-aware decisions.
In our example, with the calendar pop-up dynamic in its context, the agent correctly anticipates the outcome of its click and plans its subsequent actions accordingly, successfully navigating the semantic mismatch.
Example prompts for exploration and dynamics extraction are shown in Appendix sections~\cref{sec:webarena_prompts} and~\cref{sec:bfcl_prompts}.

\begin{wraptable}{r}{0.43\textwidth}
    \vspace{-2em}
    \caption{Number of tasks per website in the WebArena benchmark. The benchmark consists of six websites, including a multi-site category for tasks that require interacting across multiple websites (from the six sites), for a total of 812 tasks.}
    \vspace{0.5em}
    \begin{tabular}{cc}
    \hline
    \textbf{Website} & \textbf{Number of Tasks} \\ \hline
    Shopping         & 187                      \\
    Shopping Admin   & 182                      \\
    GitLab           & 180                      \\
    Map              & 109                      \\
    Reddit           & 106                      \\
    Multi-site       & 48                       \\
    Total            & 812                      \\ \hline
    \end{tabular}
    \label{tab:webarena_task_count}
    \vspace{-1.5em}
\end{wraptable}

\section{Experiments}
\subsection{Setup}
We evaluate our approach on three realistic agentic benchmarks: BFCLv3~\citep{patil_berkeley_nodate}, WebArena~\citep{zhou_webarena_2024}, and Tau-Bench~\citep{yao_tau-bench_2024}, which together cover both web navigation and conversational function calling tasks.
As for baselines, we evaluate open-source \texttt{Qwen2.5-14B-Instruct}~\citep{qwen_qwen25_2025} model and closed-source \texttt{GPT-4.1}.
On WebArena, we additionally compare to the World-model-augmented (WMA) agent from~\citep{chae_web_2025} as a baseline.
This approach first collects exploration demonstrations, then trains a \texttt{Llama-3.1-8B-Instruct}~\citep{grattafiori_llama_2024} model to predict the next state given an action (i.e., a learned world model). 
During evaluation, this world model is used to provide next-state predictions to the agent on-the-fly, helping the agent reason about the consequences of its actions.
In this comparison, we use the same \texttt{GPT-4o mini} as the original WMA experiments.
Aside from the \parametricmethodname{} and \nonparametricmethodname{} experiments, we provide ablation studies in \cref{subsec:ablation}.

For all \nonparametricmethodname{} experiments, we leverage \texttt{GPT-4.1} as an exploration policy to extract environmental dynamics. For each environment, we conduct $10$ exploration episodes to gather a diverse set of dynamics. For environment dynamics filtering, we apply \texttt{o3}\footnote{\url{https://platform.openai.com/docs/models/o3}} to remove repetitive and trivial environment dynamics. We also include ablation studies on the number of exploration episodes and the effect of filtering in \cref{subsec:ablation}.

Before delving into the results, we first introduce the benchmarks and describe the specific experimental setups used for experiments.

\paragraph{WebArena} The WebArena benchmark consists of five self-hosted websites and a total of 812 tasks. Many tasks require interacting across multiple websites, which poses additional challenges for LLM-based agents. For evaluation, WebArena employs a string-matching-based mechanism to assess correctness, providing more stable and objective results compared to benchmarks using LLMs to score, such as~\citep{he_webvoyager_2024, gou_mind2web_2025}. 
For \nonparametricmethodname{}, we first synthesize $10$ personas via prompt instruction~\ref{lst:web_persona_synthesis} for each website using GPT-4o~\citep{openai_gpt-4_2024} and with seed personas from NNetNav~\citep{murty_nnetnav_2025}. Then for each persona, we roll-out a \texttt{GPT-4.1} agent to explore the environment guided by the persona description, with the exploration prompt instruction~\ref{lst:web_exploration} derived from NNetNav, and with a maximum exploration budget of 30 steps.
To extract environment dynamics from interaction logs, we summarize each state transition pair (observation, action taken, new observation) with prompt instruction~\ref{lst:web_dynamics_extraction}, and concatenate all summarized state transition pairs to obtain the list of environmental dynamics $E=\{e\}_{i=1}^{M}$ for each website.
Finally, we utilize the \texttt{o3} model to remove repetitive and trivial environment dynamic entries from the list via prompt instruction~\ref{lst:web_filter_dynamics}, resulting in $E_{\text{clean}}$. We provide all prompts we use in~\cref{appendix}.
With \parametricmethodname, we reset the adaptation vector after each episode and we use a learning rate of 0.1 and training step of 1 for all experiments.
During evaluation, we use the text-based accessibility tree as the observation type and run experiments using BrowserGym\footnote{\url{https://github.com/ServiceNow/BrowserGym}} and AgentLab\footnote{\url{https://github.com/ServiceNow/AgentLab}} with a maximum step of $30$.

\begin{table}[]
    \centering
    \vspace{-1em}
    \caption{Main results on WebArena, WebVoyager, BFCLv3, and Tau-Bench benchmarks. We report task success rates (\%) for each model and adaptation method. For Tau-Bench, we average runs across 5 seeds and use a custom more stable codebase. Both test-time adaptation strategies improve performance. "N/A" indicates not applicable---AWM requires training a web-specific world model on state transitions, which is not applicable to Tau-Bench (conversational only) or BFCLv3 (no explicit web-like state transition data); Dynamics grounding does not operate in conversational Tau-Bench as there is no explicit and fixed state transition rules in the environment.}
    \vspace{0.5em}
    \begin{tabularx}{\textwidth}{>{\raggedright\arraybackslash}X c c c c}    \hline
    \multicolumn{1}{c}{\textbf{Model}}      & \multicolumn{4}{c}{\textbf{Task}}                         \\
                                            & WebArena    & BFCLv3      & \multicolumn{2}{c}{Tau-Bench} \\
    \multicolumn{1}{c}{}                    &             &             & Airline       & Retail        \\ \hline
    GPT-4.1                                 & 30.0        & 55.5        & -             & -             \\
    GPT-4.1 (+ \nonparametricmethodnameshort)              & 35.0 (+5.0) & 64.0 (+8.5) & N/A             & N/A             \\
    GPT-4o mini                             & 12.0        & -        & -             & -             \\
    GPT-4o mini (+ WMA)                      & 13.5 (+1.5) & N/A        & N/A             & N/A             \\
    GPT-4o mini (+ \nonparametricmethodnameshort)           & 18.0 (+6.0) & -        & N/A             & N/A             \\ \hline
    Qwen2.5-14B-Instruct                    & 17.0        & 18.5        & 21.6          & 43.3          \\
    Qwen2.5-14B-Instruct (+ \parametricmethodnameshort)     & 18.0 (+1.0) & 20.0 (+1.5) & 25.2 (+3.6)   & 44.9 (+1.6)   \\
    Qwen2.5-14B-Instruct (+ \nonparametricmethodnameshort) & 20.0 (+3.0) & 22.0 (+3.5) & N/A             & N/A             \\
    Qwen2.5-14B-Instruct (hybrid)           & 21.0 (+4.0) & 21.0 (+2.5) & N/A             & N/A             \\
    \hline
    \end{tabularx}
    \label{tab:main}
    \vspace{-1em}
\end{table}
\paragraph{BFCLv3} BFCLv3~\citep{patil_berkeley_nodate} is a benchmark with various realistic APIs to evaluate the functions calling abilities of LLM. It contains $8$ different API domains. We provide the number of APIs for each environment in~\cref{tab:bfcl_num_functions}.
We choose the multi-turn-base split to assess the agent's multi-turn function calling capabilities. This split is multi-turn and multi-step, meaning that instead of just handling a single tool call in a one-off request, models have to carry out sequences of function calls across multiple dialogue turns. BFCLv3 uses state-based evaluation (checking the system’s resulting state) and response-based evaluation (checking whether the call sequences are valid and minimally sufficient) to assess correctness, therefore being stable and deterministic.
For \nonparametricmethodname, we synthesize $10$ exploration tasks with \texttt{GPT-4o} conditioned on the function documentation for each domain via prompt instruction~\ref{lst:bfcl_exploration_synthesis}. Concretely, we instruct the LLM to think about \textit{what functions it wants to explore in order to find out unpredictable outcomes or interesting behaviors, given the function documentation}. 
We use a \texttt{GPT-4.1} agent to execute the exploration task and we collect interaction logs for environment dynamics extraction. During exploration, we ask the exploration agent to decide if the steps taken so far are sufficient to answer the exploration task with prompt instruction~\ref{lst:bfcl_exploration_user_prompt}. The environment dynamics are extracted using prompt instruction~\ref{lst:bfcl_extract_dynamics}.
We follow the same procedure as in WebArena to remove repetitive and trivial environment dynamics with prompt instruction~\ref{lst:bfcl_filter_env_dynamics}.
For \parametricmethodname, we reset adaptation vector at each turn (when a new user query arrives).


\paragraph{Tau-Bench} Tau-Bench~\citep{yao_tau-bench_2024} is a realistic function calling benchmark that simulates real-world service agents interacting with an LLM-simulated user.
It comprises two domains: airline (50 tasks) and retail (115 tasks).
The airline domain is generally more challenging than retail, as it features a much larger database, increasing the complexity of the tasks.
Tau-bench's use of an LLM-simulated user agent introduces significant variance in evaluation results, as the simulated human may drift from the original instruction or be influenced by the agent’s responses over multiple conversational turns.
To address this, we adopt the improved evaluation codebase from~\citep{prabhakar_apigen-mt_2025}, which substantially reduces reward variance and improves evaluation fidelity. Specifically, we (1) fix incorrect labels, (2) set the temperature of the LLM-simulated user agent to 0 for deterministic responses, and (3) employ a Best-of-N (N=5) sampling strategy combined with a self-critique mechanism, allowing the simulated user to better adhere to the task instruction and avoid being misled by the agent’s outputs. This protocol has been shown to improve consistency in agent performance evaluation across multiple trials according to~\citep{prabhakar_apigen-mt_2025}.
Tau-Bench consists solely of API-based tasks without observable state transitions, meaning there are no environment dynamics to extract. As a result, \nonparametricmethodname, which relies on extracting and leveraging such dynamics, is not applicable to this benchmark. Therefore, we only evaluate the \parametricmethodname\ method on Tau-Bench. To further reduce variance, we report results averaged over 5 random seeds in~\cref{tab:main}.

\subsection{Results and Analysis}
\begin{table}[]
    \centering
    \vspace{-1em}
    \caption{Success rates (\%) on the WebArena benchmark across different websites and models. Both adaptation strategies improve performance over baselines. and \nonparametricmethodname\ (\nonparametricmethodnameshort) surpasses WMA on \texttt{GPT-4o mini}. \nonparametricmethodnameshort\ improves success rate substantially on complex multi-site split.}
    \vspace{0.5em}
    
    \small{
    \begin{tabular}{>{\raggedright\arraybackslash}p{4.1cm} c c c c c c c}
    \hline
    \textbf{}                          & \multicolumn{7}{c}{\textbf{Website}}                                      \\
    \multicolumn{1}{c}{\textbf{Model}} & Gitlab & Shop Admin & Shop & Reddit & Map  & Multi & Avg \\ \hline
    GPT-4.1                             & 48.0\%   & 31.0\%           & 37.0\%     & 29.0\%   & 28.0\% & 2.0\%        & 30.0\%    \\
    GPT-4.1 (+\nonparametricmethodnameshort)                      & 45.0\%   & 27.0\%           & 33.0\%     & 42.0\%   & 31.0\% & 23.0\%       & 35.0\%    \\
    GPT-4o mini                        & 9.0\%    & 12.0\%           & 18.0\%     & 8.0\%    & 12.0\% & 12.0\%       & 12.0\%    \\
    GPT-4o mini (+WMA)                  & --    & --           & --     & --    & -- & --       & 13.5\%    \\
    GPT-4o mini (+\nonparametricmethodnameshort)                 & 18.0\%   & 14.0\%           & 26.0\%     & 14.0\%   & 19.0\% & 12.0\%       & 18.0\%    \\
    Qwen2.5-14B-Instruct               & 20.0\%   & 14.0\%           & 22.0\%     & 13.0\%   & 13.0\% & 6.0\%        & 17.0\%    \\
    Qwen2.5-14B-Instruct (+\parametricmethodnameshort)         & 16.0\%   & 14.0\%           & 24.0\%     & 25.0\%   & 16.0\% & 6.0\%        & 18.0\%    \\
    Qwen2.5-14B-Instruct (+\nonparametricmethodnameshort)        & 23.0\%   & 13.0\%           & 24.0\%     & 23.0\%   & 20.0\% & 10.0\%       & 20.0\%    \\
    Qwen2.5-14B-Instruct (Hybrid)      & 26.0\%   & 13.0\%           & 24.0\%     & 31.0\%   & 19.0\% & 4.0\%        & 21.0\%    \\
    \hline
    \end{tabular}
    }
    \label{tab:webarena}
    \vspace{-1em}
\end{table}

\paragraph{Both Adaptation Strategies Consistently Improve Performance}
As shown in~\cref{tab:main}, both \parametricmethodname{} (\parametricmethodnameshort) and \nonparametricmethodname{} (\nonparametricmethodnameshort) improve performance across models and tasks.
\nonparametricmethodname{} yields substantial improvements for \texttt{GPT-4.1}, which is better able to leverage the provided dynamics due to its superior instruction-following abilities. It is also evident in~\cref{tab:main} that our \texttt{GPT-4o mini} surpasses WMA that uses a learned world model by a large margin.
For \texttt{Qwen2.5-14B-Instruct}, both methods provide gains, highlighting the general applicability of test-time adaptation.
The improvements of \nonparametricmethodname{} are especially significant for models with stronger instruction-following abilities (e.g., GPT-4.1), likely because they can better leverage the provided dynamics context.
\begin{wraptable}{H}{0.3\linewidth}
    \centering
    \vspace{-2.2em}
    \caption{Relative latency gain (\%) of PA on BFCLv3 multi-turn. LR fixed to $0.1$.}
    \vspace{0.5em}
    \begin{tabular}{cc}
    \hline
    \textbf{Step} & \textbf{Latency Gain $\downarrow$} \\ \hline
    0             & 0\%                      \\
    1             & 3.0\%                  \\
    2             & 5.9\%                  \\
    3             & 8.2\%                  \\
    4             & 12.6\%                 \\
    5             & 15.6\%                 \\ \hline
    \end{tabular}
    \vspace{-1em}
    \label{tab:latency}
\end{wraptable}
\paragraph{Both Adaptation Strategies are computationally efficient}

The two proposed strategies are computationally efficient, though they have different profiles. \parametricmethodname{} is highly efficient for real-time use.
As shown in \cref{tab:latency}, a single update step adds only a 3\% latency overhead.
In contrast, \nonparametricmethodname{} is a one-time, deployment-time investment designed to be far more lightweight than traditional world-modeling approaches like the WMA baseline.
For example, while WMA requires synthesizing 870 tasks to collect trajectories for training a separate \texttt{Llama-3.1-8B} model, our method uses only 50 exploratory rollouts and requires no additional model training.
This streamlined process avoids WMA's costly training phase and its significant inference overhead, where action simulation takes approximately 140 seconds per step.
The total upfront cost for our \nonparametricmethodname{} procedure on a single WebArena website is approximately 7M tokens, a cost that is amortized across all subsequent tasks. 
This investment pays significant dividends where environmental dynamics are unpredictable.
For instance, on the challenging multi-site split, the success rate of our agent increases from 2\% to 23\%.
This demonstrates the value of efficiently learning an environment's dynamics at test time to compensate for the limitations of an agent's pre-training.

\paragraph{Effectiveness Depends on Environment Complexity and Method Synergy}
The smaller gains for \nonparametricmethodnameshort{} on simpler websites like ``Shopping'' suggest that when an environment's dynamics align with common sense priors (e.g., clicking 'search' shows a search bar), the explicit dynamics provide less new information, and the overhead of a longer context may slightly hinder performance.
In these simple settings, longer input on dynamics may negatively impact the model's effectiveness, as prior work has shown that excessively long contexts can hurt performance~\citep{beltagy_longformer_2020, liu_lost_2024}. We further analyze this phenomenon and provide exemplar environment dynamics in \cref{appendix:trivial_dynamics}.
Furthermore, our test of a naive hybrid approach showed mixed results, underperforming the \nonparametricmethodname{} method alone on BFCLv3 (21.0\% vs 22.0\%). 
This suggests a simple combination is insufficient and may create interference between the two adaptation signals.
A key challenge for future work is developing a more principled integration of semantic guidance and distributional alignment.

\begin{table}[]
    \centering
    \vspace{-1em}
    \caption{Ablation on exploration policy and dynamics extractor backbones on WebArena. Here we use different backbones of exploration policy and dynamics extractor. We found for \nonparametricmethodname---using the same LLM agent (itself) improves the same as using a stronger one.}
    \vspace{0.5em}
    \scalebox{0.97}{
    \begin{tabular}{lllc}
    \hline
    \multicolumn{1}{c}{\textbf{Task Model}} & \multicolumn{1}{c}{\textbf{Exploration Policy}} & \multicolumn{1}{c}{\textbf{Dynamics Extractor}} & \textbf{SR} \\ \hline
    GPT-4o mini (baseline)                   & --                                               & --                                               & 12.0        \\
    GPT-4o mini                              & GPT-4.1                                          & GPT-4.1                                          & 18.0 (+6.0) \\
    GPT-4o mini (self)           & GPT-4o mini                                      & GPT-4o mini                                      & 19.0 (+7.0) \\ \hline
    Qwen2.5-14B-Instruct (baseline)         & --                                               & --                                               & 17.0        \\
    Qwen2.5-14B-Instruct                    & GPT-4.1                                          & GPT-4.1                                          & 20.0 (+3.0) \\
    Qwen2.5-14B-Instruct (self) & Qwen2.5-14B-Instruct                            & Qwen2.5-14B-Instruct                            & 20.0 (+3.0) \\ \hline
    \end{tabular}
    }
    \vspace{-1em}
    \label{tab:ablation_exploration_policy}
\end{table}

\subsection{Ablations}
\label{subsec:ablation}


\paragraph{Dynamics grounding enables self-improvement}
We ablate the choice of backbone LLMs for exploration and dynamics extraction in our \nonparametricmethodname{} method (\cref{tab:ablation_exploration_policy}). Results show that using the same model for both components (self-improvement) performs as well as using a stronger model, indicating that our approach is robust to the choice of exploration policy and dynamics extractor.

\paragraph{Effect of filtering environmental dynamics}
We conduct an ablation experiment on BFCLv3 and find that filtering improves success rate across exploration budgets. Specifically, with 10 episodes, success rate rises from 61.0 to 64.0 (+3.0).

\paragraph{Syntactic alignment is robust to different hyperparameters}
As shown in~\cref{tab:ablation_model_sizes}, training with different learning rates generally improves performance across a range of reasonable hyperparameter values. However, using extreme hyperparameters can degrade performance. For example, in the case of the 7B model, applying a very high learning rate of 1.0 or training for 5 steps leads to reduced success rates, highlighting the importance of careful hyperparameter selection for optimal adaptation.

\paragraph{Larger models benefit from a larger learning rate} When testing with a very large learning rate (i.e., learning rate of 1.0), we find that larger models (i.e., 14B and 32B models) benefit more with a larger learning rate. We attribute this to the fact that the dimension of the adaptation vector for 14B and 32B models is the same (i.e., 5120), and is much larger compared to the 7B model's (i.e., 3584).


\section{Limitations and Future Work}
Our study is primarily focused on the Qwen2.5 model family for \parametricmethodname; future work should validate these findings across a broader range of open-source architectures.
Our \nonparametricmethodname{} method currently operates in environments with explicit state transitions, where testbeds such as user-oriented conversational tasks are not explored.  
Additionally, our online \parametricmethodname{} method does not currently normalize updates by the hidden dimension size, which could improve robustness to hyperparameters.

The most promising direction for future work is the development of a principled method for integrating our two adaptation strategies as our analysis suggests that simply combining them is suboptimal.
A more advanced approach could involve a meta-controller that assesses an environment's complexity to dynamically decide whether to rely on efficient online adaptation for simple tasks or to deploy the more costly dynamics grounding for complex, unfamiliar environments.



\section{Conclusions}

In this work, we presented a comparative study of two distinct strategies for adapting LLM agents to novel, complex environments.
We identified two primary failure modes---syntactic and semantic mismatches---and investigated targeted solutions for each.
The first, \parametricmethodname{} (\parametricmethodnameshort), is an efficient technique for aligning an agent's output distribution to the local format of an environment.
The second, \nonparametricmethodname{} (\nonparametricmethodnameshort), is an effective and efficient method for providing the agent with a causal world model through proactive dynamics extraction.

Our evaluations on diverse agentic benchmarks demonstrate that both strategies significantly improve performance.
Online adaptation provides consistent, low-cost gains, while dynamics grounding is critical for success in complex settings where an agent's pre-trained knowledge fails.
Our findings highlight that a multi-faceted approach to adaptation is necessary for robust agent generalization and point toward exciting future directions in creating more intelligent and integrated adaptation systems.


\section*{Acknowledgments}
We thank Salesforce AI Research for supporting this work. We also thank the reviewers and the area chairs for their time and thoughtful feedback.

\bibliography{references}
\bibliographystyle{iclr2026_conference}

\appendix
\section{Appendix}%
\label{appendix}
\subsection{Experiment details}
\paragraph{WebArena}
We follow the BrowserGym guide\footnote{\url{https://github.com/ServiceNow/BrowserGym/blob/main/browsergym/webarena/README.md}} to setup WebArena enviornment. We use AWS EC2 instance with a \texttt{t3a.xlarge} instance. For all exploration episodes of \nonparametricmethodname, we use a temperatue of 1.0. We use a temperature of 0.0 for all task evaluation experiments. 
\paragraph{BFCLv3}
For BFCLv3, we follow the official setup\footnote{\url{https://github.com/ShishirPatil/gorilla/tree/main/berkeley-function-call-leaderboard}} and consistently using a temperature of 0.
\paragraph{Tau-bench}
We adopt the codebase form~\citep{prabhakar_apigen-mt_2025} and use Best-of-N (N=5) strategy for LLM-simulated user agent, so each trajectory will be scored by an LLM judge and the reponse that best adheres with user intent will be select, therefore drastically mitigating variance in average reward.  
\subsection{Prompts used in WebArena experiments}
\label{sec:webarena_prompts}
\begin{lstlisting}[label=lst:web_persona_synthesis,caption=prompt used to synthesize personas for WebArena, frame=tlrb,backgroundcolor=\color{light-gray},breaklines]
You are an expert at creating realistic user personas given a website. You are given the name of a website and some example outputs for the website, generate a diverse list of ${n_personas} personas, each with a unique "persona" name and a 1-2 sentence "description" of their typical behavior, interests, or motivations on that site.

Format your output as a JSON array, where each element is an object with two fields:
- "persona": a short, descriptive name for the persona (e.g., "Casual Browser", "Tech Enthusiast")
- "description": a brief description of how this persona typically uses the website

Now, generate a list of personas in the specified JSON format. You are given the name of the website and some example outputs for the website.
Website: ${website}
Example outputs: ${examples}
\end{lstlisting}
\begin{lstlisting}[label=lst:web_exploration,caption=prompt used for persona-driven exploration in WebArena, frame=tlrb,backgroundcolor=\color{light-gray},breaklines]
You are an autonomous intelligent agent tasked with navigating a web browser.  Your objective is to simulate a task that a person might perform, by interacting with the browser through the use of specific actions.

Here's the information you'll have:

The current web page's accessibility tree: This is a simplified representation of the webpage, providing key information.
The current web page's URL: This is the page you're currently navigating.
The open tabs: These are the tabs you have open.
The previous action: This is the action you just performed. It may be helpful to track your progress.
Trajectory: This is a sequence of natural language descriptions of the agent's interaction with the web-browser.
Person Description: The description of a specific kind of person whose task you are supposed to simulate.
Environment dynamics: Descriptions of how states transition.

The actions you can perform fall into several categories:

Page Operation Actions:
`click [id]`: This action clicks on an element with a specific id on the webpage.
`type [id] [content] [press_enter_after=0|1]`: Use this to type the content into the field with id. By default, the "Enter" key is pressed after typing unless press_enter_after is set to 0.
`hover [id]`: Hover over an element with id.
`press [key_comb]`:  Simulates the pressing of a key combination on the keyboard (e.g., Ctrl+v).
`scroll [direction=down|up]`: Scroll the page up or down.

Tab Management Actions:
`new_tab`: Open a new, empty browser tab.
`tab_focus [tab_index]`: Switch the browser's focus to a specific tab using its index.
`close_tab`: Close the currently active tab.

URL Navigation Actions:
`goto [url]`: Navigate to a specific URL.
`go_back`: Navigate to the previously viewed page.
`go_forward`: Navigate to the next page (if a previous 'go_back' action was performed).

Completion Action:
`stop ["done"]`: Issue this action when you are done.

Homepage:
If you want to visit other websites, check out the homepage at http://homepage.com. It has a list of websites you can visit.

To be successful, it is very important to follow the following rules:
1. You should only issue an action that is valid given the current observation
2. You should only issue one action at a time.
3. You should follow the examples to reason step by step and then issue the next action.
4. Generate the action in the correct format. Start with a "In summary, the next action I will perform is" phrase, followed by action inside ``````. For example, "In summary, the next action I will perform is ```click [1234]```".
5. To generate an interesting task, make sure you issue atleast 4 actions before stopping. More interesting tasks typically involve more interactions with the browser.
6. You can issue atmost 40 actions before stopping, but feel free to output the stop action early if you want to stop exploring. Don't generate anything after stop.
Here are the list of environmental dynamics of this environment:

OBSERVATION:
[1744] link 'HP CB782A#ABA 640 Inkjet Fax Machine (Renewed)'
		[1749] StaticText '$279.49'
		[1757] button 'Add to Cart'
		[1760] button 'Add to Wish List'
		[1761] button 'Add to Compare'
URL: http://onestopmarket.com/office-products/office-electronics.html
ENVIRONMENTAL DYNAMICS:
[{'initial_state': 'Homepage of onestopmarket.com, with the button Add to Cart visible', 'action': 'click [1757] where the button is purchase'}]
PREVIOUS ACTION: None
Let's think step-by-step. This page lists the information of HP Inkjet Fax Machine, which is the product identified in the objective. Its price is $279.49. I have checked the environmental dynamics and see that stopping with this answer from this state has not been done before, so this action will contribute to exploration diversity. I think I have achieved the objective. In summary, the next action I will perform is ```stop [$279.49]```
OBSERVATION:
[164] textbox 'Search' focused: True required: False
[171] button 'Go'
[174] link 'Find directions between two points'
[212] heading 'Search Results'
[216] button 'Close'
URL: http://openstreetmap.org
ENVIRONMENTAL DYNAMICS:
[{'initial_state': 'OpenStreetMap homepage with search box and Go button visible', 'action': 'click [171] to submit a search'}, {'initial_state': 'OpenStreetMap homepage with search box focused', 'action': 'type [164] [parks near NY] [1]'}]
PREVIOUS ACTION: None
Let's think step-by-step. This page has a search box whose ID is [164]. According to the nominatim rule of openstreetmap, I can search for restaurants near a location by typing "restaurants near" followed by the location. I can submit my typing by pressing Enter afterwards. Before proceeding, I will check the environmental dynamics to see if a similar action has already been performed from this state. I see that there is no previous action where, starting from the current state, someone searched for "restaurants near CMU" in the search box. This means my action will contribute to exploration diversity. In summary, the next action I will perform is ```type [164] [restaurants near CMU] [1]```
OBSERVATION:
${OBSERVATION}
Person Description:
${PERSON DESCRIPTION}
URL: ${URL}
PREVIOUS ACTION: ${PREVIOUS ACTION}
ENVIRONMENTAL DYNAMICS:
${ENVIRONMENTAL DYNAMICS}

\end{lstlisting}
\begin{lstlisting}[label=lst:web_dynamics_extraction,caption=prompt used to extract environment dynamics from observations, frame=tlrb,backgroundcolor=\color{light-gray},breaklines]
You are given the output of an action taken by an autonomous intelligent agent navigating a web browser. Your objective is to produce a description of the changes made to the state of the browser. Specifically, you need to produce the following elements:
1) Initial state: provide a concise description of the browser's state before the action was taken (for example: "GitHub homepage" or "GitHub repository page").
2) Environmental dynamics: clearly describe the specific changes that occurred in the browser's state as a result of the action (for example: "The search bar expanded and is able to consume input").

Here's the information you'll have:
- **Initial state of the browser as an accessibility tree:**  
  This is a simplified representation of the webpage, providing key information.
- **Final state of the browser:**  
  This is the accessibility tree representation after the agent took the action.
- **The action taken by the web agent:**
  The agent can take actions that fall under the following categories (with descriptions):
**Page Operation Actions:**
- `click [id]`: Clicks on an element with a specific id on the webpage.
- `type [id] [content] [press_enter_after=0|1]`: Types the content into the field with id. By default, the "Enter" key is pressed after typing unless `press_enter_after` is set to 0.
- `hover [id]`: Hovers over an element with id.
- `press [key_comb]`: Simulates pressing a key combination on the keyboard (e.g., Ctrl+v).
- `scroll [direction=down|up]`: Scrolls the page up or down.

**Tab Management Actions:**
- `new_tab`: Opens a new, empty browser tab.
- `tab_focus [tab_index]`: Switches the browser's focus to a specific tab using its index.
- `close_tab`: Closes the currently active tab.

**URL Navigation Actions:**
- `goto [url]`: Navigates to a specific URL.
- `go_back`: Navigates to the previously viewed page.
- `go_forward`: Navigates to the next page (if a previous 'go_back' action was performed).

**Completion Action:**
- `stop [answer]`: Issue this action when you believe the task is complete. If the objective is to find a text-based answer, provide the answer in the bracket. If you believe the task is impossible to complete, provide the answer as "N/A" in the bracket.

**To be successful, it is very important to follow these rules:**
1. Explicitly think about the various features on the website and how the interaction with the website changed these features.
2. Provide the description of changes in one or two sentences.
3. If there is no change, your description of the changes in state should be "no change".
4. Strictly follow the JSON format in your response: {"initial_state": <description of initial state>, "environmental_dynamics": <description of environmental change>}

Here are some output examples for some random tasks:
1. {"initial_state": "Google homepage with search bar visible", "environmental_dynamics": "A modal search dialog opened up, bringing up a focused search input with suggestions, allowing the user to search or jump to various GitHub sections."}
2. {"initial_state": "Wikipedia article page", "environmental_dynamics": "The page scrolled down, revealing the 'History' section."}
3. {"initial_state": "Amazon product page", "environmental_dynamics": "A confirmation message appeared indicating the item was added to the cart."}
\end{lstlisting}
\begin{lstlisting}[label=lst:web_filter_dynamics,caption=prompt used to remove repeatitive or trivial environment dynamics in WebArena, frame=tlrb,backgroundcolor=\color{light-gray},breaklines]
You are given a list of environmental dynamics of an environment collected by an autonomous web browsing agent. You are tasked to remove repetitive and trivial environmental dynamics. Each of the point consists of the following:
1) initial_state: provide a concise description of the browser's state before the action was taken (for example: "GitHub homepage" or "GitHub repository page").
2) action: action taken to make the environmental change 
2) environmental_dynamics: clearly describe the specific changes that occurred in the browser's state as a result of the action (for example: "The search bar expanded and is able to consume input").

To be successful, it is important to follow these rules:
1. entries with trivial enviornmental dynamics, such as "when I scroll down, the page will reveal more content" should be removed. Trivial environmental dynamics are those that do not reflect a meaningful or unique change in the environment, such as simple scrolling, pagination, or toggling the visibility of already available content without introducing new information or interface elements.

2. Remove entries where the environmental_dynamics only describe the expansion or collapse of text (e.g., "The full abstract expanded and is now visible"), unless the expansion reveals new interface elements or options that were not previously accessible.

3. Remove entries where the environmental_dynamics only describe navigation to a different section of the same page (e.g., "The page scrolled to the 'References' section"), unless this navigation results in a substantial change in the available interface or content.

4. Keep entries where the environmental_dynamics describe:
   - Unexpected or spurious state change of executing an action at the initial state.
   - Keep entries where the environmental_dynamics indicate that performing an action did not result in any change to the browser's state, even though a change would normally be expected (for example, clicking a button that should open a menu but nothing happens).
   - The appearance of new controls, filters, or options in the interface.
   - Navigation to a new page or a substantially different view (e.g., from search results to a detailed article page).
   - The addition or removal of search fields, filters, or other interactive elements.
   - Any change that enables new actions or reveals new types of information not previously accessible.

5. When in doubt, prefer to remove entries that are repetitive (e.g., multiple entries describing the same type of abstract expansion) or that do not add new information about the environment's capabilities or state.

Your output should be a cleaned list of environmental dynamics, with only the non-trivial, non-repetitive, and meaningful changes retained. Please output in the original JSON format.
\end{lstlisting}
\begin{lstlisting}[label=lst:web_eval,caption=prompt used for evaluation in WebArena, frame=tlrb,backgroundcolor=\color{light-gray},breaklines]
You are an autonomous intelligent agent tasked with navigating a web browser. You will be given web-based tasks. These tasks will be accomplished through the use of specific actions you can issue.

Here's the information you'll have:
The user's objective: This is the task you're trying to complete.
The current web page's accessibility tree: This is a simplified representation of the webpage, providing key information.
The current web page's URL: This is the page you're currently navigating.
The open tabs: These are the tabs you have open.
The previous actions: These are all the action you have performed. It may be helpful to track your progress.

The actions you can perform fall into several categories:

Page Operation Actions:
`click [id]`: This action clicks on an element with a specific id on the webpage.
`type [id] [content] [press_enter_after=0|1]`: Use this to type the content into the field with id. By default, the "Enter" key is pressed after typing unless press_enter_after is set to 0.
`hover [id]`: Hover over an element with id.
`press [key_comb]`:  Simulates the pressing of a key combination on the keyboard (e.g., Ctrl+v).
`scroll [down|up]`: Scroll the page up or down.

Tab Management Actions:
`new_tab`: Open a new, empty browser tab.
`tab_focus [tab_index]`: Switch the browser's focus to a specific tab using its index.
`close_tab`: Close the currently active tab.

URL Navigation Actions:
`goto [url]`: Navigate to a specific URL.
`go_back`: Navigate to the previously viewed page.
`go_forward`: Navigate to the next page (if a previous 'go_back' action was performed).

Completion Action:
`stop [answer]`: Issue this action when you believe the task is complete. If the objective is to find a text-based answer, provide the answer in the bracket. If you believe the task is impossible to complete, provide the answer as "N/A" in the bracket.

Homepage:
If you want to visit other websites, check out the homepage at http://homepage.com. It has a list of websites you can visit.

To be successful, it is very important to follow the following rules:
1. You should only issue an action that is valid given the current observation
2. You should only issue one action at a time.
3. You should follow the examples to reason step by step and then issue the next action.
4. You are strictly forbidden from issuing a goto action to a URL that is not on the homepage.
5. Generate the action in the correct format. Start by reasoning about the current situation. End with "In summary, the next action I will perform is" phrase, followed by action inside ``````. For example, "Let's think step-by-step. Given the current state, I need to click on the like button which has id 1234. In summary, the next action I will perform is ```click [1234]```".
6. Issue stop action when you think you have achieved the objective. Don't generate anything after stop. 

Here are some example outputs for some random tasks:
1. Let's think step-by-step. This page list the information of HP Inkjet Fax Machine, which is the product identified in the objective. Its price is $279.49. I think I have achieved the objective. I will issue the stop action with the answer. In summary, the next action I will perform is ```stop [$279.49]```
2. Let's think step-by-step. This page has a search box whose ID is [164]. According to the nominatim rule of openstreetmap, I can search for the restaurants near a location by "restaurants near". I can submit my typing by pressing the Enter afterwards. In summary, the next action I will perform is ```type [164] [restaurants near CMU] [1]```
\end{lstlisting}
\begin{lstlisting}[label=lst:web_eval_with_dynamics,caption=prompt used for evaluation on WebArena with environment dynamics, frame=tlrb,backgroundcolor=\color{light-gray},breaklines]

You are an autonomous intelligent agent tasked with navigating a web browser. You will be given web-based tasks. These tasks will be accomplished through the use of specific actions you can issue.

Here's the information you'll have:
The user's objective: This is the task you're trying to complete.
The current web page's accessibility tree: This is a simplified representation of the webpage, providing key information.
The current web page's URL: This is the page you're currently navigating.
The open tabs: These are the tabs you have open.
The previous actions: These are all the action you have performed. It may be helpful to track your progress.
The environmental dynamics: You will also be given the environmental dynamics of the environment. You can use the environmental dynamics to predict the next action. Each environmental dynamics entry describes a state change in the environment and contains the following fields:
- "initial_state": The state of the environment before the action is taken.
- "action": The action performed by the agent.
- "environmental_dynamics": A description of how the environment changes as a result of the action.

The actions you can perform fall into several categories:

Page Operation Actions:
`click [id]`: This action clicks on an element with a specific id on the webpage.
`type [id] [content] [press_enter_after=0|1]`: Use this to type the content into the field with id. By default, the "Enter" key is pressed after typing unless press_enter_after is set to 0.
`hover [id]`: Hover over an element with id.
`press [key_comb]`:  Simulates the pressing of a key combination on the keyboard (e.g., Ctrl+v).
`scroll [down|up]`: Scroll the page up or down.

Tab Management Actions:
`new_tab`: Open a new, empty browser tab.
`tab_focus [tab_index]`: Switch the browser's focus to a specific tab using its index.
`close_tab`: Close the currently active tab.

URL Navigation Actions:
`goto [url]`: Navigate to a specific URL.
`go_back`: Navigate to the previously viewed page.
`go_forward`: Navigate to the next page (if a previous 'go_back' action was performed).

Completion Action:
`stop [answer]`: Issue this action when you believe the task is complete. If the objective is to find a text-based answer, provide the answer in the bracket. If you believe the task is impossible to complete, provide the answer as "N/A" in the bracket.

Homepage:
If you want to visit other websites, check out the homepage at http://homepage.com. It has a list of websites you can visit.

To be successful, it is very important to follow the following rules:
1. You should only issue an action that is valid given the current observation
2. You should only issue one action at a time.
3. You should follow the examples to reason step by step and then issue the next action.
4. You are strictly forbidden from issuing a goto action to a URL that is not on the homepage.
5. Generate the action in the correct format. Start by reasoning about the current situation. End with "In summary, the next action I will perform is" phrase, followed by action inside ``````. For example, "Let's think step-by-step. Given the current state, I need to click on the like button which has id 1234. In summary, the next action I will perform is ```click [1234]```".
6. When selecting actions, consider the environmental dynamics - what state changes will occur based on your knowledge of how the website behaves. Use this to avoid actions that would lead to undesired states or to strategically choose actions that lead to desired outcomes.
7. Issue stop action when you think you have achieved the objective. Don't generate anything after stop. 

Here are some example outputs for some random tasks:
1. Let's think step-by-step. This page lists the information of HP Inkjet Fax Machine, which is the product identified in the objective. Its price is $279.49. By considering the environmental dynamics, I can confirm that no further actions are needed to change the state, and I have achieved the objective. I will issue the stop action with the answer. In summary, the next action I will perform is ```stop [$279.49]```
2. Let's think step-by-step. This page has a search box whose ID is [164]. According to the nominatim rule of openstreetmap, I can search for the restaurants near a location by "restaurants near". By considering the environmental dynamics, I know that typing this query and pressing Enter will trigger the search and update the results accordingly. Therefore, I can submit my typing by pressing the Enter afterwards. In summary, the next action I will perform is ```type [164] [restaurants near CMU] [1]```
\end{lstlisting}
\clearpage
\subsection{Prompts used in BFCLv3 experiments}
\label{sec:bfcl_prompts}
\begin{lstlisting}[label=lst:bfcl_exploration_synthesis,caption=prompt used to synthesize exploration tasks on BFCLv3, frame=tlrb,backgroundcolor=\color{light-gray},breaklines]
You are an expert at creating exploration strategies that will help guide an LLM to explore and understand a new function calling environment. Your task is to synthesize some exploration goals based on a list of functions and a specific environment.

## Task
Given a list of functions and an environment, create ${N} distinct goals that will explore diffferent aspects of the environment. Each goal should represent different exploration strategies and approaches to discovering functionality.

## Input Format
- **Environment**: The computing environment or system (e.g., "file system", "vehicle control", "web browser")
- **Functions**: A list of available functions/commands alongside with detailed descriptions (e.g., ["cd()", "ls()", "pwd()", "mkdir()", "rm()"])

## Output Format
Output a JSON list of goals. Each goal should be a single sentence that captures the exploration goal and the strategy to achieve it.

Format: ["goal description", "goal description", ...]

## Example Output
**Environment**: File System
**Functions**:
{"name": "touch", "description": "This tool belongs to the Gorilla file system. It is a simple file system that allows users to perform basic file operations such as navigating directories, creating files and directories, reading and writing to files, etc. Tool description: Create a new file of any extension in the current directory.", "parameters": {"type": "dict", "properties": {"file_name": {"type": "string", "description": "The name of the new file in the current directory. file_name is local to the current directory and does not allow path."}}, "required": ["file_name"]}, "response": {"type": "dict", "properties": {}}}
{"name": "wc", "description": "This tool belongs to the Gorilla file system. It is a simple file system that allows users to perform basic file operations such as navigating directories, creating files and directories, reading and writing to files, etc. Tool description: Count the number of lines, words, and characters in a file of any extension from current directory.", "parameters": {"type": "dict", "properties": {"file_name": {"type": "string", "description": "Name of the file of current directory to perform wc operation on."}, "mode": {"type": "string", "description": "Mode of operation ('l' for lines, 'w' for words, 'c' for characters). ", "default": "l"}}, "required": ["file_name"]}, "response": {"type": "dict", "properties": {"count": {"type": "integer", "description": "The count of the number of lines, words, or characters in the file."}, "type": {"type": "string", "description": "The type of unit we are counting. [Enum]: [\"lines\", \"words\", \"characters\"]"}}}}
{"name": "ls", "description": "This tool belongs to the Gorilla file system. It is a simple file system that allows users to perform basic file operations such as navigating directories, creating files and directories, reading and writing to files, etc. Tool description: List the contents of the current directory.", "parameters": {"type": "dict", "properties": {"a": {"type": "boolean", "description": "Show hidden files and directories. Defaults to False. ", "default": false}}, "required": []}, "response": {"type": "dict", "properties": {"current_directory_content": {"type": "array", "description": "A list of the contents of the specified directory.", "items": {"type": "string"}}}}}
{"name": "rm", "description": "This tool belongs to the Gorilla file system. It is a simple file system that allows users to perform basic file operations such as navigating directories, creating files and directories, reading and writing to files, etc. Tool description: Remove a file or directory.", "parameters": {"type": "dict", "properties": {"file_name": {"type": "string", "description": "The name of the file or directory to remove. "}}, "required": ["file_name"]}, "response": {"type": "dict", "properties": {"result": {"type": "string", "description": "The result of the remove operation."}}}}


Generated goals:
[
  "Explore simple ways to create a file with touch(), count words of the file with wc(), and remove the file with rm().",
  "Understand the "file_name" argument of the touch() function to see if a file can be created using a path instead of just a file name."
]

## Guidelines
- **Be function-specific**: Each goal should explicitly mention which specific functions/tools will be used and in what sequence
- **Incorporate function descriptions into goal formulation**: You should examine the given function descriptions (i.e., input or output arguments) to propose goals to solve confusing or ambiguous function descriptions or environmental dynamics.
- **Ensure clarity and achievability**: Each goal should be a single, clear action that can be completed with the available functions

## Your Task
Now synthesize goal descriptions for the environment and functions provided in the format of a JSON list. Please do not leave any comment.
\end{lstlisting}
\begin{lstlisting}[label=lst:bfcl_exploration_user_prompt,caption=user prompt used in BFCLv3 exploration, frame=tlrb,backgroundcolor=\color{light-gray},breaklines]
Do you think you have well-explored all functions that you are interested in, particularly those with ambiguous input formats/structures or potentially unexpected outputs? 

Consider stopping (###STOP) if:
- You have thoroughly tested functions with unclear parameter requirements
- You have explored functions that might produce unexpected or variable output formats
- You have sufficient understanding of how to handle edge cases and errors

Continue exploring (###CONTINUE) if:
- There are still functions with ambiguous input specifications that need clarification
- You haven't fully tested functions that could have unpredictable outputs
- You need more exploration to understand parameter constraints and error handling

Please output ###STOP if you believe you have comprehensive coverage, or ###CONTINUE if more exploration is needed.
\end{lstlisting}
\begin{lstlisting}[label=lst:bfcl_extract_dynamics,caption=prompt used to extract environment dynamics from interaction logs of BFCLv3, frame=tlrb,backgroundcolor=\color{light-gray},breaklines]
You are an expert data analyst specializing in function calling environments. Your task is to analyze interaction logs between a function calling bot and an environment to extract environmental dynamics based on a given exploration goal.

## Task
Given an exploration goal and an interaction log, extract environmental dynamics that describe how the environment changes in response to bot actions. Each environmental dynamic should capture the cause-and-effect relationship between actions and environmental changes.

## Environmental Dynamic Structure
Each environmental dynamic must include these three components:

1. **initial_state**: The relevant state of the environment before the action
   - Include context like current directory, variable values, file states, or any pertinent environmental conditions
   - Be specific about what matters for understanding the action's impact

2. **action_taken**: The exact function call or action performed by the bot
   - Include the function name and all parameters with their values
   - Use the exact syntax from the interaction log

3. **environmental_dynamics**: How the environment changes as a result of the action
   - Describe what was created, modified, deleted, or revealed
   - Include any state transitions, data changes, or new information made available
   - Be specific about the observable effects

## Guidelines
- **Output format**: Generate your response as a JSON array of objects
- **Completeness**: Extract dynamics for every significant action in the log
- **Accuracy**: Base descriptions strictly on what's observable in the interaction log
- **Clarity**: Use precise, unambiguous language for each component
- **Focus**: Only include dynamics that show meaningful environmental changes

## Example Input
Goal: Investigate how to create a new file with touch() and confirm its existence using ls().

Interaction Logs:
Step 1:
Assistant: [ls()]
Environment: {"current_directory_content": []}
Step 2:
Assistant: [mkdir(dir_name="testdir")]
Environment: None
Step 3:
Assistant: [cd(folder="testdir")]
Environment: {"current_working_directory": "testdir"}
Step 4:
Assistant: [pwd()]
Environment: {"current_working_directory": "///testdir"}

## Example Output
[
  {
    "initial_state": "The current directory contains no files.",
    "action_taken": "touch(file_name=\"testfile.txt\")",
    "environmental_dynamics": "A new file named \"testfile.txt\" is created in the current directory."
  },
  {
    "initial_state": "The current directory contains the file \"testfile.txt\".",
    "action_taken": "ls()",
    "environmental_dynamics": "The environment reveals the contents of the current directory, showing that \"testfile.txt\" exists."
  }
]
\end{lstlisting}
\begin{lstlisting}[label=lst:bfcl_filter_env_dynamics,caption=prompt used to filter environment dynamics of BFCLv3, frame=tlrb,backgroundcolor=\color{light-gray},breaklines]
You are an expert data analyst specializing in function calling environments. Your task is to filter environmental dynamics to remove insignificant or environment-specific details while preserving meaningful behavioral patterns.

## Task
Given a JSON array of environmental dynamics, identify and remove dynamics that are:
1. **Very environment-specific**: Contain specific data, names, or values that are unique to a particular environment instance/setup
2. **Insignificant**: Describe trivial or non-meaningful changes that don't contribute to understanding function behavior (e.g., ls() in a file system)
3. **Redundant**: Duplicate similar patterns already captured in other dynamics

## Filtering Criteria

### Remove dynamics that contain:
- **Specific data values**: Names, IDs, file contents, user lists, or any concrete data that varies between environments
- **Environment-specific paths**: Absolute paths, specific directory names, or location-dependent information
- **Instance-specific identifiers**: User names, session IDs, timestamps, or other unique identifiers
- **Trivial state changes**: Minor formatting changes, temporary states, or cosmetic updates
- **Redundant information**: Multiple dynamics describing the same behavioral pattern with different specific values

### Keep dynamics that describe:
- **General behavioral patterns**: How functions typically behave across different environments
- **Error conditions**: Standard error messages and failure modes
- **State transitions**: General state changes (logged in/out, file created/deleted, etc.)
- **Function capabilities**: What functions can do, not what specific data they return
- **Function contains meaningful arguments**: Keep dynamics where the function call includes arguments that affect behavior or state, as these illustrate how input parameters influence outcomes.

## Guidelines
- **Output format**: Return a JSON array containing only the filtered dynamics
- **Preserve structure**: Keep the same JSON structure for remaining dynamics
- **Generalize descriptions**: Replace specific values with general descriptions where possible
- **Maintain accuracy**: Don't change the core meaning of the dynamics, only remove specificity
- **Focus on patterns**: Prioritize dynamics that show reusable behavioral patterns

## Examples

### Remove (Very environment-specific that it contains data entries specific to a environment instance):
```json
{
    "initial_state": "No user is logged into the system, and a list of users needs to be identified.",
    "action_taken": "list_users()",
    "environmental_dynamics": "Provides the list of available users: Alice, Bob, Catherine, and Daniel."
}
```

### Keep (General pattern and error message):
```json
{
    "initial_state": "No user is currently logged in.",
    "action_taken": "search_messages(keyword=\"test\")",
    "environmental_dynamics": "An error message is returned indicating that no user is currently logged in."
}
```

### Remove (Trivial transition):
```json
{
    "initial_state": "In a directory with existing files.",
    "action_taken": "ls()",
    "environmental_dynamics": "Returns a list of files and directories in the current location."
}
```

## Instructions
1. Analyze each environmental dynamic in the input JSON array
2. Apply the filtering criteria to determine if it should be kept or removed
3. For dynamics that are kept, generalize any remaining specific details
4. Return the filtered JSON array with only the significant, generalizable dynamics
\end{lstlisting}
\subsection{Additional results}
\begin{table}[h]
    \centering
    \caption{Success rate (SR) of \parametricmethodname{} across different Qwen2.5 instruct model sizes (7B, 14B, 32B) and training hyperparameters. Results are shown for varying numbers of training iterations and two learning rates (LR=0.1 and LR=1.0). Syntactic alignment adaptation generally improves SR over the baseline, with larger models and moderate learning rates yielding higher gains.}
    \vspace{0.5em}
    \begin{tabular}{cccc}
    \hline
    \textbf{Model}       & \textbf{Number of Training Iterations} & \textbf{SR (LR=0.1)} & \textbf{SR (LR=1.0)} \\ \hline
    Qwen2.5-7B-Instruct  & Baseline (0)                           & 9.00               & -                    \\
    Qwen2.5-7B-Instruct  & 1                                      & 9.50               & 9.50               \\
    Qwen2.5-7B-Instruct  & 2                                      & 9.50               & 8.50               \\
    Qwen2.5-7B-Instruct  & 3                                      & 9.50               & 9.00               \\
    Qwen2.5-7B-Instruct  & 4                                      & 10.00              & 8.50               \\
    Qwen2.5-7B-Instruct  & 5                                      & 10.00              & 8.50               \\ \hline
    Qwen2.5-14B-Instruct & Baseline (0)                           & 18.5                 & -                    \\
    Qwen2.5-14B-Instruct & 1                                      & 20.00              & 20.00              \\
    Qwen2.5-14B-Instruct & 2                                      & 20.00              & 20.00              \\
    Qwen2.5-14B-Instruct & 3                                      & 20.00              & 21.00              \\
    Qwen2.5-14B-Instruct & 4                                      & 19.50              & 20.00              \\
    Qwen2.5-14B-Instruct & 5                                      & 19.00              & 19.00              \\ \hline
    Qwen2.5-32B-Instruct & Baseline (0)                           & 26.00              & -                    \\
    Qwen2.5-32B-Instruct & 1                                      & 25.50     & 26.50              \\
    Qwen2.5-32B-Instruct & 2                                      & 26.50              & 26.50              \\
    Qwen2.5-32B-Instruct & 3                                      & 26.50              & 27.00              \\
    Qwen2.5-32B-Instruct & 4                                      & 26.50              & 27.00              \\
    Qwen2.5-32B-Instruct & 5                                      & 26.00              & 26.50
    \end{tabular}
    \label{tab:ablation_model_sizes}
\end{table}
\begin{table}[t]
    \centering
    \caption{Number of environment dynamics per website before and after filtering with \texttt{GPT-4.1} as the exploration policy and dynamics extractor. Most environment dynamics are filtered out.}
    \vspace{0.5em}
    \begin{tabular}{lcc}
    \hline
    \textbf{Website} & \textbf{Number of Env Dynamics} & \textbf{Number of Env Dynamics (after filtering)} \\ \hline
    Shopping       & 334 & 52 \\
    Shopping Admin & 334 & 50 \\
    GitLab         & 334 & 53 \\
    Map            & 334 & 18 \\
    Reddit         & 334 & 23 \\ \hline
    \end{tabular}
    \label{tab:web_num_dynamics}
\end{table}
\begin{table}[]
    \centering
    \caption{Number of functions available in each BFCLv3 environment.}
    \vspace{0.5em}
    \begin{tabular}{cc}
    \hline
    \textbf{Environment} & \textbf{Num of Functions} \\ \hline
    VehicleControlAPI    & 22                        \\
    GorillaFileSystem    & 18                        \\
    TravelAPI            & 17                        \\
    MathAPI              & 17                        \\
    TwitterAPI           & 14                        \\
    MessageAPI           & 10                        \\
    TicketAPI            & 9                         \\
    TradingBot           & 22                        \\ \hline
    \end{tabular}
    \label{tab:bfcl_num_functions}
\end{table}
\subsection{Environment dynamics analysis}
\label{appendix:trivial_dynamics}
Here we provide some concrete examples of environment dynamics that are removed by our filtering mechanism. Exploration policy and dynamics extractor is \texttt{GPT-4.1}.
\begin{itemize}
\item \textbf{No Change Dynamics}
\begin{lstlisting}[basicstyle=\ttfamily\small,breaklines=true]
{
  "initial_state": "Magento Admin product edit page, with the 'Country of Manufacture' dropdown expanded and no country selected.",
  "action": "click [1405] where [1405] is United States",
  "environmental_dynamics": "no change"
}
\end{lstlisting}
\item \textbf{Basic Dropdown/Combobox Expansion}
\begin{lstlisting}[basicstyle=\ttfamily\small,breaklines=true]
{
  "initial_state": "Magento Admin product edit page for 'Sprite Stasis Ball 65 cm', with the 'Country of Manufacture' combobox (notice-WPNSU51) collapsed and not focused.",
  "action": "click [1168] where [1168] is notice-WPNSU51",
  "environmental_dynamics": "The 'Country of Manufacture' combobox became focused and expanded, showing its list of country options."
}
\end{lstlisting}
\item \textbf{Basic Dialog Open/Close}
\begin{lstlisting}[basicstyle=\ttfamily\small,breaklines=true]
{
  "initial_state": "Magento Admin product edit page for 'Sprite Stasis Ball 65 cm', with the 'Add Attribute' button visible but no attribute selection dialog open.",
  "action": "click [720] where [720] is Add Attribute",
  "environmental_dynamics": "Clicking the 'Add Attribute' button opened the 'Add Attribute' dialog, displaying available product attributes for selection and actions such as 'Create New Attribute'."
}
\end{lstlisting}
\item \textbf{Standard Form Field Visibility}
\begin{lstlisting}[basicstyle=\ttfamily\small,breaklines=true]
{
  "initial_state": "Magento Admin product detail page for 'Sprite Stasis Ball 65 cm' with all product attribute forms visible, including a collapsed 'Country of Manufacture' dropdown (combobox).",
  "action": "click [1267] where [1267] is notice-BKW2147",
  "environmental_dynamics": "The 'Country of Manufacture' combobox was clicked, causing attribute fields for 'Country of Manufacture' and all subsequent grouped attributes to become visible."
}
\end{lstlisting}
\item \textbf{Basic Button Focus}
\begin{lstlisting}[basicstyle=\ttfamily\small,breaklines=true]
{
  "initial_state": "Magento Admin product edit page with the 'Add Attribute' dialog open; in the attributes table, the 'Options' column header includes a checkbox and an 'Options' button, neither focused nor expanded.",
  "action": "click [2372] where [2372] is ",
  "environmental_dynamics": "The 'Options' button in the 'Options' column header of the Add Attribute dialog became focused and expanded a dropdown list with a single option, 'Select All', now visible below the button."
}
\end{lstlisting}

\end{itemize}

\end{document}